\documentclass[10pt,twocolumn,letterpaper]{article}
\usepackage{cvpr}              

\usepackage{graphicx}
\usepackage{amsmath}
\usepackage{amssymb}
\usepackage{booktabs}
\usepackage{enumitem}

\usepackage[symbol]{footmisc}

\usepackage{float}
\usepackage{algorithm}
\usepackage{algpseudocode}

\usepackage[pagebackref,breaklinks,colorlinks]{hyperref}

\usepackage[capitalize]{cleveref}
\crefname{section}{Sec.}{Secs.}
\Crefname{section}{Section}{Sections}
\Crefname{table}{Table}{Tables}
\crefname{table}{Tab.}{Tabs.}

\def\cvprPaperID{*****} 
\def\confName{CVPR}
\def\confYear{2023}

\begin{document}
\title{Harnessing the Power of Text-image Contrastive Models for Automatic Detection of Online Misinformation}


\author{Hao Chen$^1$, \ Peng Zheng$^1$, \ Xin Wang$^{2*}$, \ Shu Hu$^3$, \ Bin Zhu$^4$, \ Jinrong Hu$^1$\thanks{corresponding author}, \ Xi Wu$^1$, \ Siwei Lyu$^2$\\
$^1$Chengdu University of Information Technology, China \ \ \ 
$^2$University at Buffalo, USA\\
$^3$Carnegie Mellon University, USA \ \ $^4$Microsoft, USA\\
{\tt\small \{haochen,hjr,wuxi\}@cuit.edu.cn} \quad {\tt\small \{xwang264,siweilyu\}@buffalo.edu}\\
{\tt\small 3210607015@stu.cuit.edu.cn} \quad {\tt\small shuhu@cmu.edu} \quad {\tt\small binzhu@microsoft.com}
}

\maketitle

\begin{abstract}
   As growing usage of social media websites in the recent decades, the amount of news articles spreading online rapidly, resulting in an unprecedented scale of potentially fraudulent information. Although a plenty of studies have applied the supervised machine learning approaches to detect such content, the lack of gold standard training data has hindered the development. Analysing the single data format, either fake text description or fake image, is the mainstream direction for the current research. However, the misinformation in real-world scenario is commonly formed as a text-image pair where the news article/news title is described as text content, and usually followed by the related image. Given the strong ability of learning features without labelled data, contrastive learning, as a self-learning approach, has emerged and achieved success on the computer vision. In this paper, our goal is to explore the constrastive learning in the domain of misinformation identification. We developed a self-learning model and carried out the comprehensive experiments on a public data set named COSMOS. Comparing to the baseline classifier, our model shows the superior performance of non-matched image-text pair detection (approximately 10\%) when the training data is insufficient. In addition, we observed the stability for contrsative learning and suggested the use of it offers large reductions in the number of training data, whilst maintaining comparable classification results.
\end{abstract}

\section{Introduction}
The proliferation of news articles on social media platforms allows for real-time access to information, but also leads to an increase in the spread of misinformation due to deceptive practices. Misinformation has an adverse impact on both cyber and physical societies, and has gained considerable coverage in the last several years. For example, the COVID-19 pandemic has provided a fertile ground for conspiracy theories. One of the most prominent ones was that the 5G technology was somehow responsible for the emergence of the novel coronavirus. This theory gained particular momentum in early April 2020 and led to a wave of vandalism targeting communication infrastructure in the UK and other nations~\cite{41}. In addition, misinformation causes anxiety in populations across different ages \cite{10}. 

\begin{figure}[t] 
\centering
\includegraphics[scale=0.26]{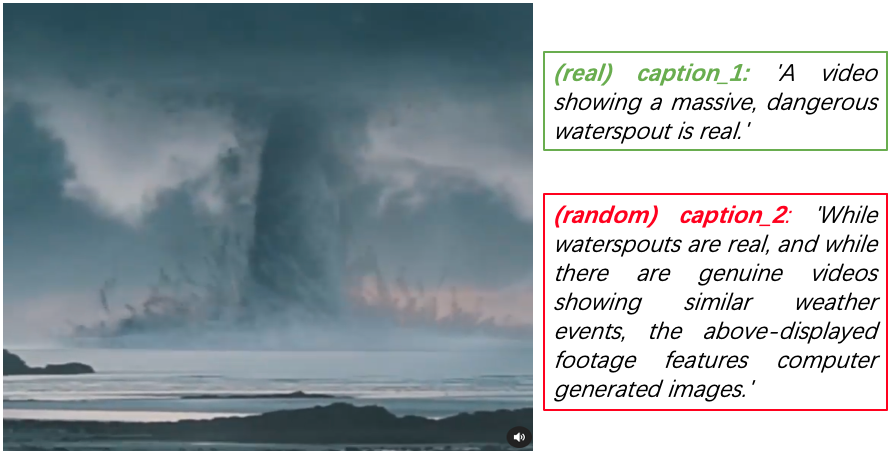}
\caption{An example of text-image pairing. For the image on the left, the text in the green box is the \textbf{real} (matched) caption, while the text in the red box is a \textbf{fake} (random) caption.}
\label{example_fig}
\vspace{-2mm}
\end{figure}

Therefore, there is an urgent demand for tools that can detect fake news content accurately and efficiently in order to eliminate misinformation and protect the harmony of online environment. Manually monitoring authenticity of information requires substantial human efforts and time \cite{wang2022gan}. In recent years, dominant automatic approaches \cite{pu2022learning, guo2022open,guo2022eyes,guo2022robust,wang2022gan,hu2021exposing} rely on machine learning techniques, particularly supervised classification models, to identify fake content. However, labelled data sets that supervised learning needs are very limited in both sizes and diversities, which presents a major obstacle for applications of supervised learning algorithms in this domain. 
Determining whether a news article is fake or not is generally a challenging task since it requires comprehensive knowledge and verification by authentic sources. 

Additionally, most researchers concentrates on using natural language processing (NLP) techniques to identify fake text-based content, overlooking the fact that news articles often contain both text and images. An example is shown in Figure.~\ref{example_fig}. The caption in the green box is the real description of the image while the caption in the red box is false. Such misinformation is easily manipulated or created by an AI-based generator and then rapidly circulated through the Internet. Relying on textual semantic and syntactic similarities \cite{9,21,23} can achieve promising identification performance to some extent. However, these models ignore relationships across multiple data modals, particularly the image-text, which can deteriorate the accuracy of identification. To address the limitations of these models, Aneja et al.~\cite{1} attempt to solve the problem by combing both data types. Specifically, their dataset, COSMOS, consists of images and two captions for each image, and their task is to predict if the two captions are both corresponding to the image, i.e., the out-of-context (OOC) classification.
They use a convolutional neural network as an image encoder and a pre-trained language model as a textual caption encoder, and achieve 85\% classification accuracy on COSMOS. However, the model is trained on the large corpus which is less efficient and time-consuming. In addition, results mostly are influenced by the pre-trained model (SBERT) according to our investigation.


In this paper, we extend Aneja et al.'s method~\cite{1} by using a language-vision model based on constrastive learning \cite{29} for out-of-context detection on the COSMOS dataset. As a self-learning approach, the constrastive learning shows a strong ability to learn feature representations without annotating a large-scale dataset. It learns representations of data by contrasting similar and dissimilar pairs of examples. The merit of contrastive learning lies in its ability to leverage the inherent structure of data to learn more useful representations. By contrasting similar examples and pulling them closer together in representation space, and pushing dissimilar examples further apart, contrastive learning can learn more robust and discriminative features that can be used for a variety of downstream tasks, such as image classification or language understanding. This has been shown to be especially effective in settings where labeled data is scarce or expensive to obtain, as contrastive learning can learn from large amounts of unlabeled data. 
We use Aneja et al.'s~\cite{1} method as the baseline in our experimental evaluation. Our results indicate that our proposed method outperforms the baseline on the out-of-context prediction. 
The main contributions of this paper can be summarized as follows:
\begin{itemize}[noitemsep,nolistsep]
    \item Our study on the baseline model reveals for the first time that the baseline relies primarily on the textual similarity from the pre-trained model (SBERT) to classify OOC, which may potentially introduce bias and distort the results; 
    \item We present a new model incorporating a contrastive learning component, which we will show is advantageous in capturing image feature representations, particularly when the training data is limited in size;
    \item We conduct extensive experiments to evaluate our proposed method and compare with the baseline method. These experiments demonstrate that our method outperforms the baseline steadily in a varying training data sizes.
    \item We developed the classification model to identify the correct caption.
\end{itemize}

We note that this work focuses on the same prediction task as the baseline on the COSMOS dataset. It is not the real-world misinformation detection wherein the prediction is to predict if a pair of image and its caption is consistent or not. This is because we do not have a dataset that can mimic the real-world misinformation detection. That said, we should point out that our proposed method with minor modifications can be applied to the scenario of the real-world misinformation detection. Actually, most of our proposed method is general and can be applied to other classification tasks. In particular, our use of contrastive learning is beneficial for scenarios when there is a lack of labeled training data. 

The paper is organized as follows: Section~\ref{sec:2} reviews related work, and Section~\ref{sec:3} descibes our proposed method. Our experimental evaluation is presented in Section~\ref{sec:4}. The conclusion and future work are presented in Section~\ref{sec:5}.

\section{Related Work}
\label{sec:2}

Online misinformation has become a topic of interests over the past few years, motivating the research community to address the problem. Bondielli et al. \cite{11} categorise information as fake or rumours depending on whether the news has been confirmed by the authoritative sources. Both Guo et al.~\cite{8} and Meel et al.~\cite{16} elaborate the differences of various terms related to misinformation on social media, such as hoax, disinformation, and fake news. 
Instead of nitpicking on nuances of different definitions, we would like to focus on machine learning techniques themselves, specifically contrastive learning. Therefore, we decide to use `out-of-context' to refer to 
misinformation comprising inconsistent image-text pairs. In this context, we briefly review related work, which includes misinformation detection and contrastive learning.

\subsection{Misinformation Detection}
Early research on the automatic identification of misinformation on social media concentrates on single-type data, particularly textual content. Traditional supervised classification techniques have been widely used in this area, such as support vector machine (SVM) \cite{13,14,hu2022rank}, na\"ive bayes \cite{12}, logistic regression \cite{18}, and decision tree \cite{14,19}. Classic feature representations such as bag of words and n-grams with TF-IDF are generally used, with semantic or syntactic information ignored 
due to individually treated features (word tokens). This issue is subsequently alleviated by using feature engineering, a process of extracting and adding both linguistic and handcrafted features manually. For example, Pelrine et al.\cite{9} systematically compare a set of transformer-based models on textual misinformation detection across various social media data sets. They 
point out the benefit of feature engineering. Shu et al. \cite{12} investigate social networks and use spatiotemporal information such as numbers of retweets, timestamp, and locations to improve classifier results. Kwon et al. \cite{17} explore users' profiles to increase the detection accuracy. However, feature engineering requires human efforts, in particular the knowledge of linguistic and social science. In addition, Zhu et al. \cite{3} point out entities in news articles can change over time, which adversely impacts detection results. Inspired by the emerge of word2vec and paragraph2vec, a plenty of recent research \cite{20,21,22,23,24} explore distributed representations where text content is converted into a dense vector by a language-embedding model, which is usually pre-trained on a general language corpus and thus preserves intrinsic language features such as syntactic and semantic features.

In addition to textual misinformation, another widely spread form of online news misinformation is de-contextualization (aka. out-of-context pairing) where images and their associated texts are 
unrelated to each other. 
Many researchers apply multi-modal analysis to tackle the problem of detecting this type of misinformation.
Singhal et al. \cite{25} introduce an ensemble model that exploits both textual and visual features to identify image-text fake news. Likewise, Giachanou et al. \cite{15} combine image features extracted by a VGG \cite{34} model and text features by a Bert \cite{35} model to detect image-text misinformation. Recently, Aneja et al.\cite{1} focus on the ``cheapfake" generated by using AI-free approaches, such as filtering, re-contextualizing, and photo-shopping, rather than on ``deepfake" generated by using AI-based techniques. They suggest utilizing image and text embeddings to forecast if an image-caption pair is out-of-context. They also release a substantial dataset for further research on this matter and it is becoming a well-known dataset in the media forensic domain. Nonetheless, their model was trained on the entire dataset. This paper aims to build upon their research by examining the efficiency of the contrastive learning model when training data is restricted.

\subsection{Contrastive Learning}
When using machine learning techniques for classification tasks, such as out-of-context (OOC) detection, data needs to be converted to a compact feature. Over the past decade, the dominant approach for determining image features is learning in a supervised way, such as training from ImageNet \cite{27}. ViT \cite{31} is a widely used visual feature representation model that uses the transformer \cite{4} framework and is trained on classification tasks. Inspired by the achievement of BERT \cite{35} in the NLP domain, computer vision community starts to increasingly focus on unsupervised training. Contrastive learning, as a self-learning approach, has gained popularity because it is able to learn feature representation without annotated data. Contrastive learning aims to move augmented samples generated from the same sample close to each other while keeping samples from different data far away. 


Many contrastive learning models have been proposed \cite{lin2022contrastive}. He et al. \cite{28} propose Momentum Contrast (MoCo) for unsupervised visual representation learning that matches encoded data with a serious of keys using the contrastive loss \cite{26}. Subsequently, Chen et al. \cite{29} propose SimCLR that generates training instances by separating different data augmentations. They comprehensively analyze a variety of image manipulation methods such as crop, resize, flip, color distort, rotate, cutout, and Gaussian noise. Gao et al. \cite{7} propose SimCSE by adapting SimCLR to textual data  that generates positive instances by different Bert dropouts and takes other samples within the batch as negative instances. Caron et al. \cite{33} propose SwAV (Swapping Assignments between Views) by modifying SimCLR that clusters data and leverages contrastive learning techniques without requiring the computation of individual augmented samples. Radford et al. \cite{5} introduce CLIP (Contrastive Language-Image Pre-training) that connects an image with a text description and creates a feature space for both data types. The model provides an efficient way to generate a text based on an image or vice versa. He et al. \cite{6} create a masked autoencoders (MAE) to reconstruct masked patches when an image is split into multiple patches. Instead of training a model by using both positive and negative samples as aforementioned methods do, Grill et al. \cite{30} propose BYOL that depends only on positive samples and is able to achieve outstanding performance for the feature representation.

\begin{figure*}[t] 
\centering
\includegraphics[scale=0.478]{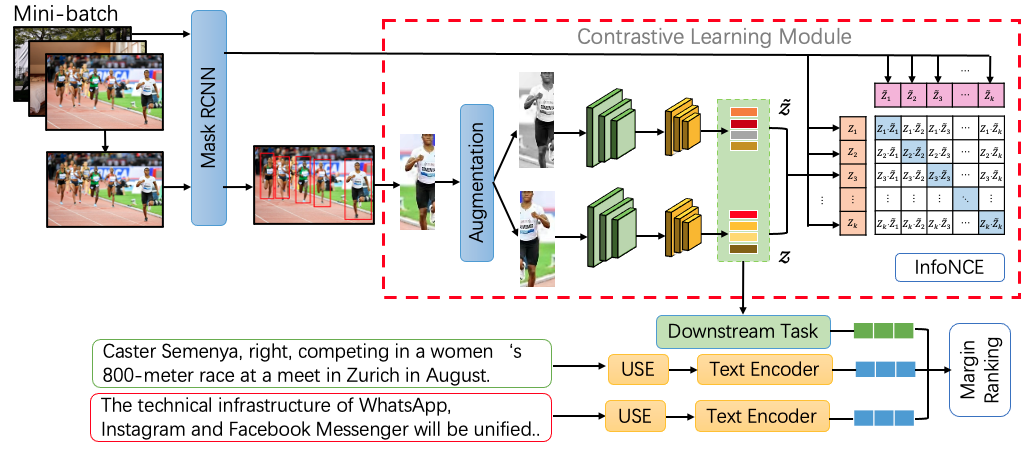}
\caption{The structure of multi-modal contrastive learning. The model is accomplished by two branches, image feature learning and classification training. First, a picture in mini-batch (the size is 64) has been fed into a Mask RCNN by which the 10 objects have been detected. Each object is then augmented individually, followed by the full connected layer to produce the dense vector denoted as $\tilde{z}$ and $z$. Subsequently, the matrix of the mini-batch is developed for the InfoNCE loss training, where the pair of $z$ and $\tilde{z}$ from the same object is treated as positive instances (the diagonal of the matrix), otherwise, the rest pairs are negative instances (see Eq. \ref{eq:CL_loss}). In the classification training, the two captions – matched (green) and another caption sampled randomly (red) – are encoded using the Universal Sentence Encoder model (USE)\cite{38}. The output of text encoder combining with the output of image encoder are passed to compute the similarities between object-caption pairs and finally used to reduce the margin ranking loss as following Eq. \ref{eq:MR_loss}.}
\label{model_fig}
\end{figure*}

\section{Our Proposed Method}
\label{sec:3}

Our goal is to leverage the power of contrastive learning in the feature representation to identify inconsistent text-image pairing on social media. We describe its training and testing details in this section.

\subsection{Contrastive Learning-guided Image-text  Matching Training}
Inspired by the baseline model on COSMOS \cite{32}, we extract features from images and texts separately and interact them to learn their matching. The training procedure is shown in Figure~\ref{model_fig} and described in detail as follows. 

\smallskip
\noindent
\textbf{Contrastive Learning Module}. For each image, we use pre-trained Masked-RCNN \cite{40} as the object detection \cite{hu2022pseudoprop} backbone to detect objects included in the image. Then we feed images and their detected objects (bounding boxes) into the augmentation module.

Within the augmentation module, each detected object is augmented. Augmented images are then fed to an image encoder followed by a full connected layer to generate a dense vector. We consider all augmented images from the same sample as positive instances and randomly selected images from other samples of the dataset as negative instances for training the constrastive learning model.
Specifically, applying Mask RCNN on the input image, we can obtain N detected objects to form a set $\{x_k\}_{k=1}^N$ of objects, where $x_k\in \mathbb{R}^{d_x}$ and $d_x$ is the dimension of detected object $x$. Then we apply data augmentation twice to get $2N$ objects, $\{\widetilde{x}_l\}_{l=1}^{2N}$, where $\widetilde{x}_{2k}$ and $\widetilde{x}_{2k-1}$ are two random augmentations of $x_k (k=1,\cdots,N)$. Different augmentation strategies can be used, such as rotations, adding noise, translation, brightness, etc. Thus an object can be augmented to generate more than two augmented objects, but only two augmented objects are used for each detected object in our setting. We use the following notation for two related augmented objects. 
Let $i\in I:=\{1,...,2N\}$ be the index of an arbitrary augmented object and $j(i)$ is the other augmented object that shares the same source object as the $i$-th augmented object. We feed augmented objects to an object encoder, represented as $E(\cdot):\mathbb{R}^{d_x}\rightarrow \mathbb{R}^{d}$ ($d$ is the output dimension of $E$), which is a ResNet-50 backbone followed by three components: RoIAlign, average pooling, and two fully-connected (FC) layers. Then, we can obtain a 300-dimensional vector for each augmented object, which maps the object feature representation into the application space of the contrastive loss, i.e., 
$\widetilde{z}_{i}=E(\widetilde{x}_{i})$ and $\widetilde{z}_{j(i)}=E(\widetilde{x}_{j(i)})$ for two augmented objects from the same source object.

\begin{algorithm}
\caption{The Out-of-Context Matching}
\label{alg_matching}
\begin{algorithmic}
\State \textbf{Data:} sample $\{ X_{k} \}^{N}_{k=1}$ in batch size $N$
\State $X_{k}=<caption_r>image<caption_m>$
\State $\tau$ and $\gamma$ are constant
\ForAll {$k\in \{ 1,...,N \}$}
  \State $O_{m} \leftarrow $ MaskRCNN$(X_{k})$ \Comment 10 object detection
  \State $C_{km} \leftarrow t \cdot (X_{km})$ \Comment{text encoding for matched}
  \State $C_{kr} \leftarrow t \cdot (X_{kr})$
  \Comment{text encoding for random}
  \ForAll{$m \in \{ 1,...,10 \}$}
    \State $\{ A,\tilde{A} \} \leftarrow A(O_{m})$ \Comment augmentation
    \State $Z \leftarrow f \cdot (A)$ \Comment image encoding
    \State $\tilde{Z} \leftarrow f \cdot (\tilde{A})$
  \EndFor
\EndFor

\State $M=N*10$ \Comment{\#of augmentation in batch}
\ForAll{$i\in \{1,...,2M\}$ and $j\in \{1,...,2M\}$}
\State $s_{i,j}=z_{i}z_{j}/(\left\| z_{i} \right\| \left\| z_{j} \right\|)$ \Comment{pairwise similarity}
\EndFor
\State \textbf{define} $\ell(i,j)$ \textbf{as} $\ell(i,j)=-\log\frac{exp(s_{i,j}/\tau)}{\sum_{k=1}^{2M}1_{[k\neq i]}exp(s_{i,k}/\tau)}$
\State $\mathcal{L}_{CL}=\frac{1}{2M}\sum_{k=1}^{{}M}\left[ \ell(2k-1,2k)+\ell(2k,2k-1) \right]$
\State \textbf{define} $s_m = \max({Z}_k \cdot {C}_{km}|k\in \{ 1,...,N \})$
\State \textbf{define} $s_r = \max({Z}_k \cdot {C}_{kr}|k\in \{ 1,...,N \})$
\State $\mathcal{L}_{Match}=[s_r-s_m+\gamma]_+$
\State update networks $f$ and $t$ to minimize $\mathcal{L}_{CL}$ and $\mathcal{L}_{Match}$ 

\end{algorithmic}
\end{algorithm}

To shorten the distance between encoder vectors $\widetilde{z}_{i}$ and $\widetilde{z}_{j(i)}$ from the same source object and widen the distance between $\widetilde{z}_{i}$ and an augmented object from another source object, we use the self-supervised contrastive learning to formulate the self-supervised  contrastive loss $\mathcal{L}_{CL}$ as follows,
\begin{equation}
    \begin{aligned}
    \mathcal{L}_{CL}=\frac{-1}{|I|}\sum_{i\in I}\log \frac{\exp(\widetilde{z}_{i} \cdot \widetilde{z}_{j(i)}/\tau)}{\sum_{a\in A(i)} \exp(\widetilde{z}_{i} \cdot \widetilde{z}_{a}/\tau)},
    \end{aligned}
\label{eq:CL_loss}
\end{equation}
where $|I|$ is the cardinality of $I$, $\tau\in \mathbb{R}^+$ is a positive scalar temperature parameter, $\cdot$ is the inner (dot) product operator, and $A(i):=I \setminus \{i\}$. It is common to regard $i$ as an anchor. $j(i)$ is called the \textit{positive} and the other $2N-2$ indices ($\{k\in A(i)\setminus\{j(i)\}\}$) are called the \textit{negatives}. The numerator in the log function of Eq.~(\ref{eq:CL_loss}) is the representation distance between $\widetilde{z}_{i}$ and $\widetilde{z}_{j(i)}$. The denominator is the representation distance between $\widetilde{z}_{i}$ and a total of $2N-1$ terms, including the positive and negatives. With this contrastive learning module, we can enhance the accuracy of representations from the encoder.

\smallskip
\noindent
\textbf{Image-text Matching Module}. This module match an image and its text (i.e., caption). Given matched caption $c_{m}$ of the input image and a random caption $c_{r}$ from a different image in the dataset,  we follow \cite{32} to use a pre-trained transformer-based Universal Sentence Encoder (USE, $U(\cdot)$) \cite{38} to encode captions into unified 512-dimensional vectors. The vectors are then sent to an additional text encoder ($T(\cdot)$) to convert to a specific dimensional feature space $\mathbb{R}^d$ that matches the the output dimension (i.e., $d$) of $E(\cdot)$. In particular, the text encoder is a ReLu followed by one FC layer. Therefore, we can represent $\widetilde{c}_m=T(U(c_m))$ and $\widetilde{c}_r=T(U(c_r))$ for the final embedded features of $c_m$ and $c_r$, respectively. 

Then we evaluate the match performance of the object embedding and the caption embedding. Specifically, we use dot product to calculate the similarity between $\widetilde{z}_i$ and $\widetilde{c}_m$ (or $\widetilde{c}_r$). We extract the maximum value as the final similarity score, 
\begin{equation*}
    \begin{aligned}
    &s_m = \max(\{\widetilde{z}_i^\top \widetilde{c}_m|i\in I\}),\\
    &s_r = \max(\{\widetilde{z}_i^\top \widetilde{c}_r|i\in I\}),
    \end{aligned}
    \label{eq:MR_loss}
\end{equation*}
where $s_m$ is the final similarity score for the matched caption and $s_r$ is the final similarity score for the random caption. Our goal is to make $s_m$ as larger as possible and $s_r$ as smaller as possible. Thus, we design the following max-margin loss for training the model,
\begin{equation}
    \begin{aligned}
    \mathcal{L}_{Match}=[s_r-s_m+\gamma]_+,
    \end{aligned}
\label{eq:match}
\end{equation}
where $[a]_+=\max(0,a)$ is the hinge function. $\gamma \in \mathbb{R}$ is a preset margin hyperparameter. The algorithm is showed in Algorithm~\ref{alg_matching}.

\subsection{Image-text Mismatching Training}


\smallskip
\noindent
\textbf{Cross Training}. We first train object encoder $E$ in the contrastive learning module based on $\mathcal{L}_{CL}$ (Eq.~(\ref{eq:CL_loss})) for all images in the dataset. 
Then we fix the contrastive learning module and train text encoder $T$ according to $\mathcal{L}_{Match}$ (Eq.~(\ref{eq:match})) on all images. The weights of the whole model are updated iteratively.

\smallskip
\noindent
\textbf{Joint Training}. In addition to cross training, we explore joint training as well. Rather than freezing one of the loss functions during training, we normalize the loss of contrastive learning module $\mathcal{L}_{CL}$ (Eq.~(\ref{eq:CL_loss})) and add $\mathcal{L}_{Match}$ (Eq. (\ref{eq:match})) to it to get the overall average loss on all images.

\begin{figure}[t]
\centering
\includegraphics[scale=0.22]{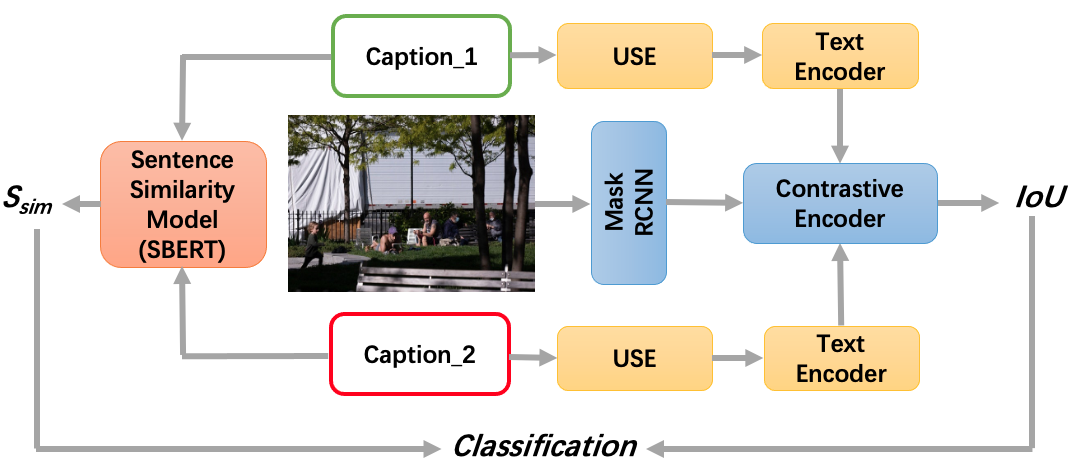}
\caption{The testing structure. $IoU$ indicates whether the two captions are describing the same object and $S_{sim}$ represents the semantic similarity between the two captions. It predicts out-of-content if both scores are higher than their preset thresholds.}
\label{testing_fig}
\end{figure}

\subsection{Image-text Mismatching Prediction}
We follow \cite{32} for our model's prediction of mismatching image and text or not, 
as shown in Figure~\ref{testing_fig}. The prediction is determined by two scores, the $IoU$ score and the Sentence BERT (SBERT) score ($S_{sim}$). The former score indicates whether the two captions are describing the same image region (object), and the latter is calculated for their sentence similarity.

Specifically, given a testing data (an image and two captions, e.g. \emph{\textless caption\_1\textgreater image \textless caption\_2\textgreater}), we use the state-of-the-art SBERT model \cite{39}, which is pre-trained on the Sentence Textual Similarity (STS) task, to get the ($S_{sim}$) score for assessing both semantic and syntactic similarities between two the sentences. SBERT takes two pieces of text content as input and output a score in the range from 0 to 1. A higher score indicates that the two captions share more similar context.

For the $IoU$ score, we use both image encoder and text encoder that are obtained from the trained language-vision model. We first compute the visual correspondences of objects $B_{IC_{i}}$ in the image for each caption respectively. For example, $B_{IC_{1}}$ represents the largest value (object) of image-caption alignment for $I$ and $caption\_1$. Then we calculate $IoU$ for the overlapping of two bounding box (area) corresponding to $caption\_1$ and $caption\_2$. At last, we apply the following rule to predict the image-caption pair is out-of-context only if the two captions are describing the same object while they are having the different semantic meaning:
\begin{itemize}
    \item Out-of-Content, if $IoU(B_{IC_{1}},$ $B_{IC_{2}})>threshold$ and $S_{sim}<threshold$; 
    \item Non Out-of-Context, otherwise.
\end{itemize}

\section{Experimental Evaluation}
\label{sec:4}

\subsection{Datasets \& Pre-processing}
As aforementioned, the lack of gold-standard training data is a major obstacle for us to investigate our approaches. To avoid this issue, this paper decided to use the data set that is large-scale and publicly available. Aneja et al. \cite{1} create the data from two primary sources, fact-checking website and various mainstream news platforms such as New York Times, CNN, Reuters and so on. The original data is documented as JSON-format file. The structure of data is formed as \emph{\textless caption\_1\textgreater image \textless caption\_2\textgreater} where each image is followed by two captions, the one is genuine and other one is synthetic. The data summary is showing as Table \ref{tab_dataset}.

We acknowledged that there is no labelling process in the training set and validation set as the synthetic caption for each image is randomly chose from the rest of text descriptions. However, the text-image pairs in testing set are manually labelled by the authors. An example of data instance is showing in Figure~\ref{example_fig}. 

For the text pre-processing, we carried out the entity extraction to replace all the names, locations, and dates to the unique tokens respectively. For instance, the caption ``Caster Semenya, right, competing in a women's 800-meter race at a meet in Zurich in August." is changed to ``Caster GPE, right, competing in a women's QUANTITY race at a meet in GPE in DATE."

\subsection{Experiments Setup}

This section will elaborate the experiments setup. Overall, we have conducted two experiments to demonstrate the benefit of contrastive learning when tackling small size of training set, and also a new model for identifying which of caption is true. To be specific, we compared three different models in our research:

\begin{itemize}
    \item \textbf{\textit{baseline}}, the model is originated from the paper \cite{32}.
    \item \textbf{\textit{cross training}}, we use the contrastive learning structure to generate the feature representation for image as shown in Figure~\ref{model_fig}. Since the model have two loss functions (\textit{InfoNCE and MarginRanking}), we iteratively freeze the one of them and train on the other.
    \item \textbf{\textit{joint training}}, we further replaced the contrastive learning model to the joint training where the two loss functions (\textit{InfoNCE and MarginRanking} are normalized and reduced at same time). 
\end{itemize}

\begin{table}[t] 
\centering
\scalebox{0.89}{
\begin{tabular}{cccc}
\hline
 & \textbf{\#of Images} & \textbf{\#of Captions} & \textbf{Annotation} \\ \hline
\textbf{Training Data} & 160k & 360k & no \\
\textbf{Validation Data} & 40k & 90k & no \\
\textbf{Testing Data} & 1700 & 1700 & yes \\ \hline
\end{tabular}}
\caption{The statistic summary of data sets}
\label{tab_dataset}
\end{table}

\noindent\textbf{Evaluation Metric.} Given the ultimate goal of this paper is to boost the ability of detecting OOC content, we use the standard classification evaluation metrics (\textit{Accuracy}, \textit{Precision}, \textit{Recall} and \textit{F1-score}).




\noindent\textbf{Implementation Details.} In addition, we acknowledge that detecting OOC is a trade-off issue where we have to decide whether the cost of false negatives (OOC has been classified as non-OOC) is higher than the false positives (non-OOC has been classified as OOC). We assume that in a real-world scenario, failing to identify misinformation and allowing it to spread would have a greater impact on social networks than incorrectly labeling clean content as misinformation. In this context, we will extend the original research that is only showing accuracy, and put more focus on the recall (also known as true positive rate). 

Considering the neural network is randomly initialized, we carried out our experiments three times and get the averages. In addition, we used the following hyper-parameters as default in proposed model: Adam optimizer, ReLU activation function, a batch size of 64, and training for 10 epochs. 

\subsection{Contrastive Learning vs. Baseline}
\label{sec:4-3}
Rather than using the full size of training data, we would like to investigate the performance of aforementioned models on the small training set. We used the configurations as following, \textit{Mask-RCNN}: 10 objects are detected; \textit{Augmentation}: we chose one of noises such as rotation, adding gray, filtering, resizing, translation, brightness, etc.; \textit{Resnet}: 18 convolutional layers are implemented and the output is 512 dimension (also text encoder is 512 dimension); \textit{Dense Layer}: the output of dense vector is 300 dimension.

The results have been presented in Table \ref{result_tab}, highlighting the best performance for each evaluation metric. It is observed that the baseline model performs the worst across all four metrics. As anticipated, replacing the structure of the original convolutional network with a contrastive learning module in the image encoder improves the ability to detect OOC content. Both cross training and joint training models achieve an accuracy of 0.80, which is an improvement of approximately 10\% over the baseline model (0.73). In terms of recall, the contrastive learning models achieve better results than the baseline as well, with cross training and joint training models achieving 0.9 and 0.92 respectively. The increased performance is attributed to the enlargement of the training samples through augmentation, indicating that the contrastive learning models can effectively address the issue of insufficient training data. However, based on the paired t-test, the difference between cross training and joint training contrastive models is not statistically significant (p-value is 0.718).

\begin{table}[t]
\centering
\scalebox{0.882}{
\begin{tabular}{ccccc}
\hline
 & \textit{\textbf{Accuracy}} & \textit{\textbf{Precision}} & \textit{\textbf{Recall}} & \textit{\textbf{F1-Score}} \\ \hline
\textit{\textbf{Baseline}} & 0.73 & 0.74 & 0.87 & 0.80 \\
\textit{\textbf{Cross Training}} & \textbf{0.80} & \textbf{0.75} & 0.9 & \textbf{0.82} \\
\textit{\textbf{Joint Training}} & \textbf{0.80} & 0.74 & \textbf{0.92} & \textbf{0.82} \\ \hline
\end{tabular}}
\caption{The comparison results of three models over the Accuracy, Precision, Recall and F1-score. }
\label{result_tab}
\end{table}

We recognize that the accuracy of the aforementioned methods is lower than the results reported in the original paper, primarily because we randomly selected a subset of the training data, which consisted of less than 5,000 samples. In comparison to the reported accuracy of 85\% achieved by using the full training set of 16,000 samples, we demonstrate that contrastive learning can achieve nearly 94\% (0.80/0.85) of the performance using only 28\% (4.5k/16k) of the training data. Furthermore, adding more data for training would yield limited improvements.

\subsection{Contrastive Learning for True Caption Classification}
We identified that most of classification decision is based on the sentence similarity BERT model which is pre-trained without fine-tuned. As shown in Figure~\ref{testing_fig}, the final prediction is decided by the $IoU$ and $S_{sim}$ scores. However, the majority of testing data (more than 80\%) has nearly 0.9 of $IoU$ which is above the threshold (0.5). Consequently, the final classification is mainly attributes to the $S_{sim}$. To alleviate the bias, we would like to directly use contrastive in the training phase. 

\begin{figure}[t]
\centering
\includegraphics[scale=0.262]{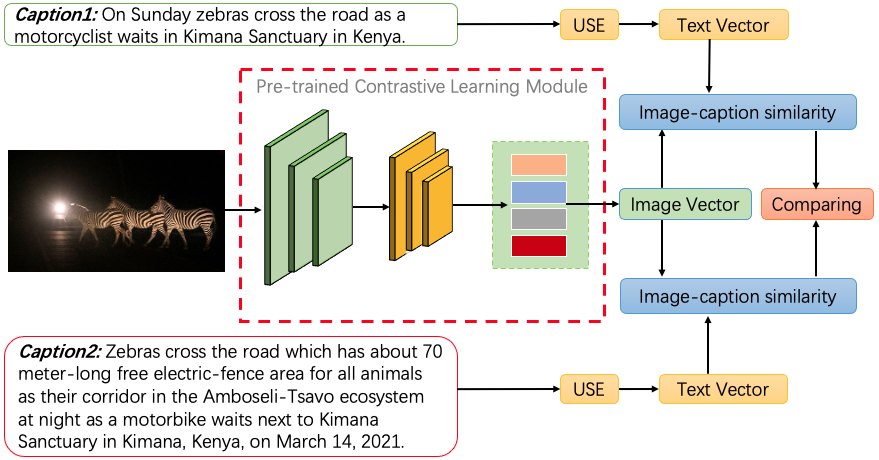}
\caption{Classification of true caption based on the image-caption model. The pre-trained contrastive learning module (see Figure~\ref{model_fig}) is used for the image feature representation, and two captions are encoded through the Universal Sentence Encoder respectively. Then the caption vectors are used to compute the similarity with image vector. Finally, the prediction is decided by the similarity score where the image vector is closer to the true caption than the false one.}
\label{classification_fig}
\end{figure}

We simply modify our model for the task of classifying the correct caption. The structure is illustrated in Figure~\ref{classification_fig}. Firstly, we extract the vector representation for the image from the trained contrastive learning model. Subsequently, we calculate the cosine similarity between this representation and the vector representations of the two captions (matched and non-matched captions). Finally, we determine the true caption based on the similarity score where the matched caption should have higher value with the image. The results obtained are mediocre, as only 941 out of 1700 (55\%) have been correctly detected. We will continue working on improving this method in the future work.

\subsection{Comparison on Varying Training Data Sizes}

\begin{figure*}[t]
\centering
\includegraphics[scale=0.352]{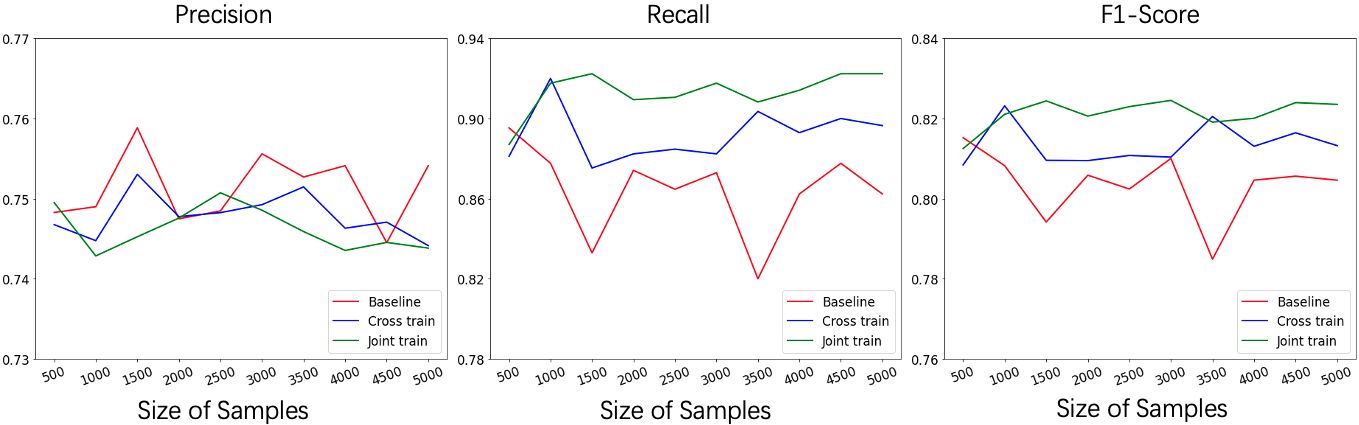}
\caption{OOC classification precision, recall and F1-score for three different models with varying size of training samples.}
\label{fullresults_fig}
\end{figure*}

\begin{figure}[t]
\centering
\includegraphics[scale=0.3]{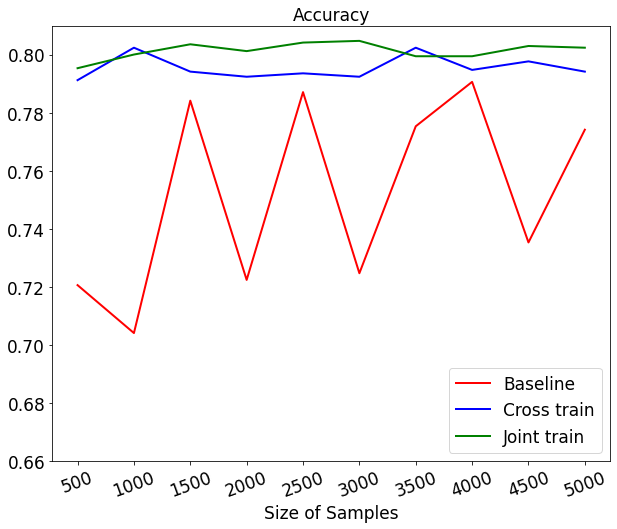}
\caption{OOC classification accuracy for three different models with varying size of training samples.}
\label{results_fig}
\end{figure}

Following the results we obtained from previous experiment (Section \ref{sec:4-3}), we would like further to investigate the impact of varying levels of training size on performance of the models. To be detailed, we want to examine whether the contractive learning would achieve comparable results when training on the even smaller size of data, and how the various mount of training data would consequent the models' classification ability. To do this, we created 10 different levels of training sizes, ranging from 500 to 5000 samples in intervals of 500 (randomly selected from original data set). For each level, we use the identical training data across three models. Then, we measured classification accuracy on the testing set as the same way with the previous experiment. The results are shown in Figure~\ref{results_fig}.

Theoretically, increasing the training data should improve the models' classification performance. However, according to the Figure~\ref{results_fig}, the baseline model showed the opposite trend. For instance, the best accuracy (0.78) of red line was achieved at 1500 training samples, and the baseline model is surprisingly decreased multiple times at training over the 2000, 3000, and 4500 samples. All three drops were significant, with the accuracy decreasing from nearly 0.78 to 0.72. Although the baseline models were fluctuating, the overall trend was upward (red line increased from 0.72 to 0.77). 

In contrast, both contrastive learning models achieved superior results (around 0.79) when trained on very few data samples. For instance, the results of the two contrastive models fell in the range of 0.78 to 0.80, while the lowest accuracy of the baseline model was 0.72. Moreover, their performance steadily improved with the addition of more data, although the improvement was limited. 

Overall, the two contrastive learning models performed comparably, and it was challenging to determine which one was better. To further explore this, we also reported the precision, recall, and f1-score in Figure~\ref{fullresults_fig}. The joint training achieved the best results in terms of recall and f1-score, followed by the cross training, while the baseline model had the worst performance.


\section{Conclusions}
\label{sec:5}
The focus of this work was to investigate the performance of constrastive learning for the feature representation when the tackling the domain of labelled data insufficiency, specifically the text-image context pairing. We highlight the following points from this work: (1) We proposed an advanced model for the task of out-of-context (OOC) detection based on the contrastive learning which is a self-supervised machine learning technique and utilize data augmentation for training. According to the results, we demonstrated the superior of our model comparing to the benchmark from original paper; (2) We focused on the situation where the labelled data is inadequate for training, which is a general limitation for the most of classification tasks. We have done the comparison and show that the contrastive learning has a strong ability to learn the image feature and achieve the 94\% of full performance, albeit decreasing the training data size nearly 70\%; (3) We carried out a comprehensive analysis of the ability of the different classifiers to deal with the varying training data sizes. Our results show that the contrastive learning model produce the stable performance and increase the accuracy steadily once adding the training data. However, the baseline model has ups-and-downs results.

At last but not the least, we notice that one of the disadvantages for the COSMOS data set is that the label of OOC is based on whether the two captions are consistent with each other and also corresponding to the image. In this case, it ignores the real scenario of judging which of the caption is true. Although we proposed a new classifier model to deal with this issue, the results are not promising. 

In the future, we would like further to explore contrastive learning and focus on the advance techniques to improve the misinformation detection accuracy.

{\small
\bibliographystyle{ieee_fullname}
\bibliography{egbib}
}

\end{document}